\documentclass[a4paper,twoside]{article}

\usepackage{epsfig}
\usepackage{subcaption}
\usepackage{calc}
\usepackage{amssymb}
\usepackage{amstext}
\usepackage{amsmath}
\usepackage{amsthm}
\usepackage{multicol}
\usepackage{pslatex}
\usepackage{apalike}
\usepackage{algorithm}
\usepackage{algorithmic}
\usepackage{dsfont}
\usepackage[bottom]{footmisc}
\usepackage{SCITEPRESS}     % Please add other packages that you may need BEFORE the SCITEPRESS.sty package.

\newtheorem{definition}{Definition}

\begin{document}

\title{Swarm Behavior Cloning}

\author{\authorname{Jonas Nüßlein\sup{1}, Maximilian Zorn\sup{1}, Philipp Altmann\sup{1} and Claudia Linnhoff-Popien\sup{1}}
\affiliation{\sup{1}Institute of Computer Science, LMU Munich, Germany}
\email{jonas.nuesslein@ifi.lmu.de}
}


\abstract{
In sequential decision-making environments, the primary approaches for training agents are Reinforcement Learning (RL) and Imitation Learning (IL). Unlike RL, which relies on modeling a reward function, IL leverages expert demonstrations, where an expert policy $\pi_e$ (e.g., a human) provides the desired behavior. Formally, a dataset $D$ of state-action pairs is provided: $D = {(s, a = \pi_e(s))}$. A common technique within IL is Behavior Cloning (BC), where a policy $\pi(s) = a$ is learned through supervised learning on $D$. Further improvements can be achieved by using an ensemble of $N$ individually trained BC policies, denoted as $E = {\{\pi_i(s)\}}_{1 \leq i \leq N}$. The ensemble’s action $a$ for a given state $s$ is the aggregated output of the $N$ actions: $a = \frac{1}{N} \sum_{i} \pi_i(s)$.
This paper addresses the issue of increasing action differences—the observation that discrepancies between the $N$ predicted actions grow in states that are underrepresented in the training data. Large action differences can result in suboptimal aggregated actions. To address this, we propose a method that fosters greater alignment among the policies while preserving the diversity of their computations. This approach reduces action differences and ensures that the ensemble retains its inherent strengths, such as robustness and varied decision-making.
We evaluate our approach across eight diverse environments, demonstrating a notable decrease in action differences and significant improvements in overall performance, as measured by mean episode returns.
}

\onecolumn \maketitle \normalsize \setcounter{footnote}{0} \vfill

\section{\uppercase{Introduction}}

Reinforcement Learning (RL) is a widely recognized approach for training agents to exhibit desired behaviors by interacting with an environment over time. In RL, an agent receives a state $s$, selects an action $a$, and receives feedback in the form of a reward, learning which behaviors are beneficial (high reward) and which are detrimental (low reward) based on the reward function provided by the environment \cite{sutton2018reinforcement}. While RL has shown great promise, designing an effective reward function can be highly challenging. A well-designed reward function must satisfy several criteria: (1) the optimal behavior should yield the maximum possible return $R^*$ (the sum of all rewards in an episode); (2) suboptimal behaviors must be penalized, resulting in a return $R < R^*$, ensuring that shortcuts or unintended strategies are discouraged; (3) the reward function should be dense, providing informative feedback at every step of an episode rather than just at the end; (4) the reward should support gradual improvement, avoiding overly sparse rewards such as those that assign 1 to the optimal trajectory and 0 to all others, which can hinder exploration and learning \cite{eschmann2021reward,knox2023reward}.

Due to the inherent complexity of crafting such reward functions, the field of \textit{Imitation Learning} (IL) has emerged as an alternative approach \cite{zheng2021imitation,torabi2019recent}. Instead of relying on an explicitly defined reward function, IL uses expert demonstrations to model the desired behavior. This paradigm has proven effective in various real-world applications, such as autonomous driving \cite{bojarski2016end,codevilla2019exploring} and robotics \cite{giusti2015machine,finn2016guided}. A prominent method within IL is Behavior Cloning (BC), where supervised learning is applied to a dataset of state-action pairs $D = \{(s, a = \pi_e(s))\}$, provided by an expert policy $\pi_e$. Compared to other IL methods like Inverse Reinforcement Learning \cite{zhifei2012survey,nusslein2022case} or Adversarial Imitation Learning \cite{ho2016generative,torabi2019adversarial}, BC has the advantage of not requiring further interactions with the environment, making it particularly suitable for non-simulated, real-world scenarios.

A straightforward extension of BC is the use of an ensemble of $N$ individually trained policies. In this approach, the \textit{ensemble action} is computed by aggregating the $N$ predicted actions as $a = \frac{1}{N} \sum_{i} \pi_i(s)$. Although ensemble methods often improve robustness, they can encounter challenges when states in the training data $D$ are underrepresented. For these states, the $N$ policies may predict actions $\{a_i\}_{1 \leq i \leq N}$ that diverge significantly, leading to suboptimal aggregated actions.

\begin{figure*}[htp]
\centering
\minipage{1\textwidth}
  \centering
  \includegraphics[width=\linewidth]{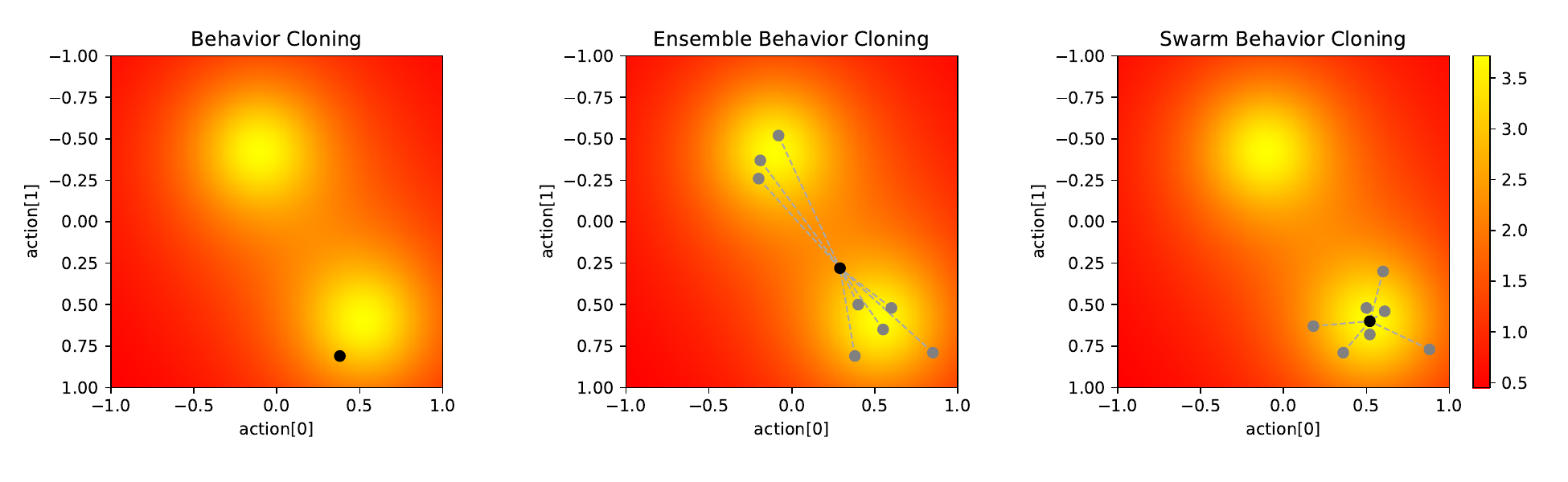}
  \caption{This figure visualizes schematically the predicted actions of three different Behavior Cloning approaches, represented as black dots, in a 2-dim action space for some state $s_t$. The heatmap represents the Q-values $Q(a_t, s_t)$. (Left) the left plot shows plain \textit{Behavior Cloning}. A policy $\pi$ was trained using supervised learning on some training data $D$. The black dot is the predicted action $a_t = \pi(s_t)$. (Middle) in \textit{Ensemble Behavior Cloning} an ensemble of $N$ policies is trained individually on $D$. The $N$ predicted actions $\{a_i = \pi_i(s_t)\}$ (gray dots) are then aggregated to the ensemble action (black dot). (Right) in our approach \textit{Swarm Behavior Cloning} an ensemble of $N$ policies is trained as well. However, they are not trained individually but using a modified loss function, see formula (2). The effect is a smaller difference between the $N$ predicted actions, resembling a swarm behavior. Similar to \textit{Ensemble Behavior Cloning} the ensemble action (black dot) is then aggregated from the $N$ predicted actions (gray dots).}
\endminipage
\end{figure*}

In this paper, we address the problem of \textit{increasing action differences} in such underrepresented states. Specifically, we propose a new loss function that encourages greater alignment among the $N$ policies in the ensemble while preserving the diversity of their computations. This approach reduces action differences and ensures that the ensemble retains its inherent strengths, such as robustness and varied decision-making. As illustrated in Figure 1, this approach—termed \textit{Swarm Behavior Cloning}—leads to more consistent predictions across the ensemble, with the $N$ predicted actions (gray dots) clustered more closely together compared to standard Ensemble Behavior Cloning (middle plot). By minimizing action divergence, our approach improves the quality of the aggregated action and enhances performance in diverse environments.

\section{\uppercase{Background}}

\subsection{Reinforcement Learning}
Reinforcement Learning (RL) problems are often modeled as Markov Decision Processes (MDP). An MDP is represented as a tuple $M = \langle S, A, T, r, p_0, \gamma \rangle$ where $S$ is a set of states, $A$ is a set of actions, and $T(s_{t+1}\:|\:s_t,a_t)$ is the probability density function (pdf) for sampling the next state $s_{t+1}$ after executing action $a_t$ in state $s_t$. It fulfills the Markov property since this pdf solely depends on the current state $s_t$ and not on a history of past states $s_{\tau < t}$. $r : S \times A \rightarrow \mathbb{R}$ is the reward function, $p_0$ is the start state distribution, and $\gamma \in [0;1)$ is a discount factor which weights later rewards less than earlier rewards \cite{phan2023attention}.
\par
A deterministic \textit{policy} $\pi : S \rightarrow A$ is a mapping from states to actions. Return $R = \sum_{t=0}^{\infty} {\gamma}^{\;t} \cdot r(s_t,a_t)$ is the (discounted) sum of all rewards within an episode. The task of RL is to learn a policy such that the expected cumulative return is maximized:

\begin{equation*}
\pi^* = \underset{\pi}{\mathrm{argmax}} \:\: J_{p_0}(\pi, M) = \underset{\pi}{\mathrm{argmax}} \:\: \mathbb{E} \: \Biggl[\sum_{t=0}^{\infty} {\gamma}^{\;t} \cdot r(s_t,a_t) \: | \: \pi \Biggr]
\end{equation*}
\ \\
\ \\
Actions $a_t$ are selected following policy $\pi$. In Deep Reinforcement Learning the policy $\pi$ is represented by a neural network $\hat{f_{\phi}}(s)$ with a set of trainable parameters $\phi$ \cite{sutton2018reinforcement}.

\subsection{Imitation Learning}

Imitation Learning (IL) operates within the framework of Markov Decision Processes, similar to Reinforcement Learning (RL). However, unlike RL, IL does not rely on a predefined reward function. Instead, the agent learns from a dataset of expert demonstrations consisting of state-action pairs:
\[
D = \{(s_i, a_i = \pi_e(s_i))\}_i
\]
where each $a_i$ represents the expert’s action $\pi_e(s_i)$ in state $s_i$. IL is particularly useful in situations where demonstrating the desired behavior is easier than designing a corresponding reward function.

IL can be broadly divided into two main categories: Behavior Cloning (BC) and Inverse Reinforcement Learning (IRL). Behavior Cloning focuses on directly mimicking the expert’s actions by training a policy through supervised learning on the provided dataset $D$ \cite{torabi2018behavioral}. In contrast, Inverse Reinforcement Learning seeks to infer the underlying reward function $r_e(s, a)$ that would make the expert’s behavior optimal, using the same dataset $D$.

The key advantage of BC over IRL is that BC does not require further interactions with the environment during training. This makes BC more applicable to real-world (non-simulated) scenarios, where collecting new trajectories can be costly, time-consuming, or even dangerous due to the exploratory actions involved \cite{zheng2021imitation,torabi2019recent}.

In addition to BC and IRL, there are adversarial approaches to IL that do not neatly fit into these two categories. These methods also necessitate environment rollouts for training. The core idea behind adversarial methods is to frame the learning process as a zero-sum game between the agent and a discriminator. The discriminator’s objective is to distinguish between state-action pairs generated by the agent and those produced by the expert, while the agent tries to generate actions that fool the discriminator \cite{ho2016generative,torabi2019adversarial}.

\begin{figure*}[h]
\centering
\minipage{1\textwidth}
  \centering
  \includegraphics[width=\linewidth]{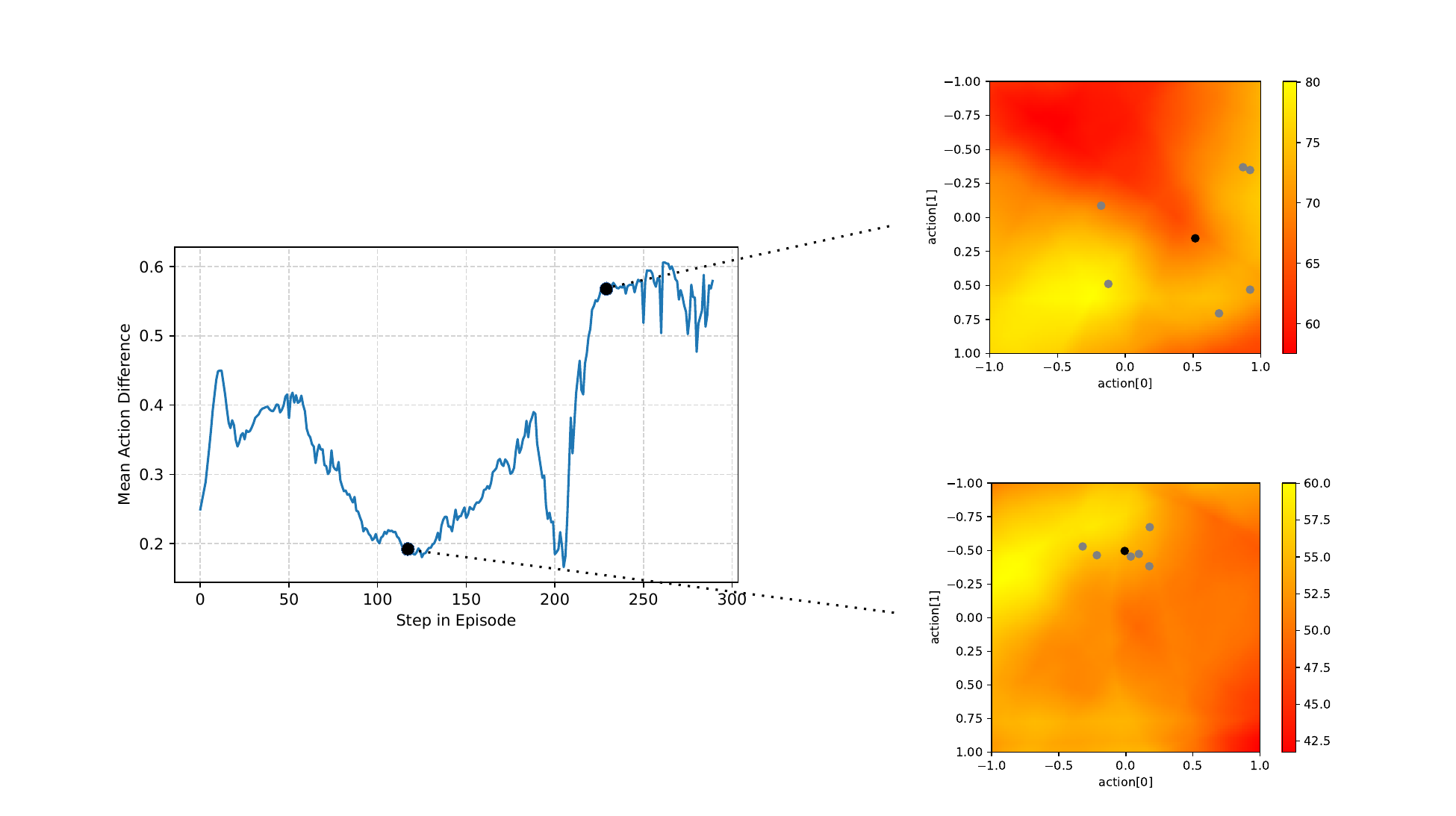}
  \caption{This figure visualizes exemplarily the \textit{mean action difference} for an entire episode of an ensemble containing $N = 6$ policies. We used the \textit{LunarLander-continuous} environment since it has a 2-dim action space that can be easily visualized. The x-axis in the left plot represents the timestep in the episode. For two interesting timesteps, we have visualized the predicted actions of the $N$ policies $\{a^i_t = \pi_i(s_t)\}$ (gray dots) as well as the aggregated action (black dot) on a 2-dim map (the complete action space). The underlying heatmap represents the Q-values from the expert critic (a fully-trained SAC model from Stable-Baselines 3).}
\endminipage
\end{figure*}

\section{\uppercase{Problem Analysis: Action Difference in Ensemble Behavior Cloning}}

When training an ensemble of $N$ policies, denoted as $\{\pi_i\}_{1 \leq i \leq N}$, on a given dataset $D = \{(s_t, a_t)\}_t$ consisting of state-action pairs, each policy $\pi_i$ is trained independently to predict actions based on the input states. Due to differences in the training paths and the inherent variability in the learning process, the ensemble members typically produce different action predictions for the same state $s$. This divergence can be quantified through the concept of \textit{mean action difference}, which we formally introduce in Definition 3.1.

\begin{definition}[Mean Action Difference]
Let $E = \{\pi_i\}_{1 \leq i \leq N}$ represent an ensemble of $N$ policies, and let $s$ be a given state. For the ensemble $E$, we can compute $N$ action predictions, denoted as $A = \{a_i = \pi_i(s)\}_{1 \leq i \leq N}$. The \textit{mean action difference}, $d$, is defined as the average pairwise L2-norm between the actions in $A$. Formally, the mean action difference $d$ is given by:
\[
d = \frac{2}{N(N-1)} \sum_{i} \sum_{j>i} \lVert a_i - a_j \rVert
\]
This measure quantifies the average difference between the action predictions of the ensemble members for a particular state $s$. A higher value of $d$ indicates greater divergence in the actions predicted by the policies, whereas a lower value suggests that the ensemble members are more consistent in their predictions.
\end{definition}

In Figure 2 (left), we illustrate the \textit{mean action difference} across an entire episode within the \textit{LunarLander-continuous} environment. The ensemble in this experiment consisted of six policies trained on a dataset $D$ derived from a single expert demonstration. The expert used for this demonstration was a fully-trained Soft-Actor-Critic (SAC) model from Stable-Baselines 3 \cite{stable-baselines3}. The figure highlights areas with both high and low mean action differences. In regions with high \textit{mean action difference}, the ensemble members produce actions that are quite divergent, whereas regions with low \textit{mean action difference} show greater agreement among the predictions.

To explore this phenomenon further, we analyzed two specific timesteps, $t = 120$ and $t = 225$, where we visualized the actions predicted by the $N = 6$ policies (represented by gray dots) alongside the aggregated action (black dot) on the 2D action space of this environment. The Q-value heatmap, provided by the expert SAC critic network, is also displayed. Notably, in the upper-right plot, we observe a scenario where the Q-value of the aggregated action, $Q(a, s)$, is lower than the Q-values of all individual actions, i.e., $Q(a, s) < Q(a_i, s)$ for all $i$. This phenomenon is more prevalent in states where the \textit{mean action difference} is large, leading to suboptimal aggregated actions due to the inconsistency among the policies.

In Chapter 5, we present a modified training loss designed to tackle the problem of divergent action predictions within the ensemble. This new loss function encourages the individual policies in the ensemble to learn more similar hidden feature representations, effectively reducing the \textit{mean action difference}. \textbf{By fostering greater alignment among the policies while preserving the diversity of their computations, the ensemble retains its inherent strengths—such as robustness and varied decision-making—while producing more consistent actions}. This reduction in action divergence improves the quality of the aggregated actions, ultimately leading to enhanced overall performance.

\section{\uppercase{Related Work}}

Imitation Learning is broadly divided into Behavior Cloning (BC) and Inverse Reinforcement Learning (IRL) \cite{zheng2021imitation,torabi2019recent}. While in Behavior Cloning the goal is to learn a direct mapping from states to actions \cite{bain1995framework,torabi2018behavioral,florence2022implicit}, IRL is a two-step process. First, the missing Markov Decision Process (MDP) reward function is reconstructed from expert data, and then a policy is learned with it using classical reinforcement learning \cite{arora2021survey,ng2000algorithms}. Besides BC and IRL, there are also adversarial methods such as GAIL \cite{ho2016generative} or GAIfO \cite{torabi2019adversarial}. BC approaches cannot be adequately compared to IRL or adversarial methods, since the latter two require access to the environment for sampling additional episodes. Therefore, we compare our approach only against other Behavior Cloning approaches: BC \cite{bain1995framework} and Ensemble BC \cite{yang2022towards}.

Besides these approaches, other Behavior Cloning algorithms exist which, however, require additional inputs or assumptions. In \cite{brantley2019disagreement} the algorithm \textit{Disagreement-regularized imitation learning} is presented that first learns a standard BC ensemble. In the second phase, another policy is learned by using a combination of BC and RL, in which the reward function is to not drift into states where the variance of the ensemble action predictions is large. The policy therefore has two goals: (1) it should act similarly to the expert (2) the policy should only perform actions that ensure the agent doesn't leave the expert's state distribution. However, this approach also requires further interactions with the environment making it inappropriate to compare against a pure Behavior Cloning algorithm. The similarity to our approach \textit{Swarm BC} is that both algorithms try to learn a policy that has a low \textit{mean action difference}. Our algorithm can therefore be understood as a completely offline version of \textit{Disagreement-regularized imitation learning}.
\ \\
\ \\
In \cite{smith2023strong} the proposed algorithm \textit{ILBRL} requires an additional exploration dataset beyond the expert dataset and subsequently uses any Offline RL algorithm. In \cite{hussein2021robust} a data-cleaning mechanism is presented to remove sub-optimal (adversarial) demonstrations from $D$ before applying BC. \cite{torabi2018behavioral} proposes a state-only BC approach using a learned inverse dynamics model for inferring the executed action.

In \cite{shafiullah2022behavior} Behavior Transformers are introduced for learning offline from multi-modal data. The authors in \cite{wen2020fighting} present a BC adaption for combating the "copycat problem" that emerges if the policy has access to a sliding window of past observations. In this paper, we assume a Markov policy that only receives the last state as input. Therefore the "copycat problem" does not apply here.

There is much literature on ensemble methods \cite{dong2020survey,sagi2018ensemble,dietterich2002ensemble,zhou2021ensemble,webb2004multistrategy}. The main difference of current ensemble methods to our approach is that we encourage the ensemble members to reduce the output diversity, while current methods try to increase the output diversity. We show that in Markov Decision Problem environments, ensembles with large action diversities can lead to poor aggregated actions. In the next chapter, we therefore present an algorithm that reduces the \textit{mean action difference}.

\section{\uppercase{Swarm Behavior Cloning}}

In this section, we introduce our proposed approach, \textit{Swarm Behavior Cloning} (Swarm BC), which aims to reduce the divergence in action predictions among ensemble policies by encouraging them to learn similar hidden feature representations.

We assume that each of the $N$ policies in the ensemble $E = {\{\pi_i\}}_{1 \leq i \leq N}$ is modeled as a standard Multilayer Perceptron (MLP). The hidden feature activations of policy $\pi_i$ at hidden layer $k$, given input state $s$, are represented as $h_{ik}(s) \in \mathbb{R}^m$, where $m$ is the number of neurons in that hidden layer. These hidden activations form the basis of the ensemble's predictions.

Consider a training data point $(s,a) \in D$, where $s$ is the state and $a$ is the expert's action. In standard Behavior Cloning (BC), each policy in the ensemble is trained individually using a supervised learning loss function. The goal is to minimize the difference between each policy’s predicted action $\pi_i(s)$ and the corresponding expert action $a$. The standard loss for training a BC ensemble is given by:

\begin{equation} L(s,a) = \sum_{i} \left( \pi_i(s) - a \right)^2 \end{equation}

This formulation treats each policy independently, which can lead to divergence in their predicted actions, especially in underrepresented states, resulting in a high \textit{mean action difference}.

\begin{algorithm}[tb]
    \caption{Swarm Behavior Cloning}
    \label{alg:algorithm}
    \textbf{Input}: expert data $D=\{(s, a = \pi_e(s))\}$\\
    \textbf{Parameters}: \\ $\tau$ (regularization coefficient) \\ $N$ (number of policies in the ensemble)\\
    \textbf{Output}: trained ensemble $E = \{\pi_i\}_{1 \leq i \leq N}$. Predict an action $a$ for a state $s$ using formula (3)
    \begin{algorithmic}[1] %[1] enables line numbers
        \STATE initialize $N$ policies $E = \{\pi_i\}_{1 \leq i \leq N}$
        \STATE train ensemble $E$ on $D$ using loss (2)
        \STATE \textbf{return} trained ensemble $E$
    \end{algorithmic}
\end{algorithm}

The core idea behind \textit{Swarm BC} is to introduce an additional mechanism that encourages the policies to learn more similar hidden feature activations, which in turn reduces the variance in their predicted actions. This is achieved by modifying the standard loss function to include a regularization term that penalizes large differences in hidden feature activations between policies. The adjusted loss function is:

\begin{equation} L(s,a) = \sum_{i} \left( \pi_i(s) - a \right)^2 + \tau \sum_{k} \sum_{i<j} \left( h_{ik}(s) - h_{jk}(s) \right)^2 \end{equation}

The first term is the standard supervised learning loss, which minimizes the difference between the predicted action $\pi_i(s)$ and the expert action $a$. The second term introduces a penalty for dissimilarity between the hidden feature activations of different policies at each hidden layer $k$. The hyperparameter $\tau$ controls the balance between these two objectives: reducing action divergence and maintaining accuracy in reproducing the expert’s behavior.

\textbf{By incorporating this regularization term, the individual policies in the ensemble are encouraged to align their internal representations of the state space, thereby reducing the \textit{mean action difference}. At the same time, the diversity of the ensemble is preserved to some extent, allowing the individual policies to explore different solution paths while producing more consistent outputs.}

The final action of the ensemble, known as the \textit{ensemble action}, is computed as the average of the actions predicted by the $N$ policies, following the standard approach in ensemble BC:

\begin{equation} a = \frac{1}{N} \sum_{i} \pi_i(s) \end{equation}

This averaging mechanism allows the ensemble to benefit from the collective knowledge of all policies, while the regularization ensures that the predictions remain aligned.

The overall procedure for \textit{Swarm BC} is summarized in Algorithm 1. The algorithm takes as input the expert dataset $D = {(s, a = \pi_e(s))}$, the number of ensemble members $N$, and the hyperparameter $\tau$. It outputs a trained ensemble $E = {\{\pi_i\}}_{1 \leq i \leq N}$ capable of making robust action predictions by computing the ensemble action as shown in Equation (3).

\section{\uppercase{Experiments}}

In this experiments section, we want to verify the following two hypotheses:

\begin{itemize}
\item Using our algorithm \textit{Swarm Behavior Cloning} we can reduce the \textit{mean action difference} as defined in \textbf{Definition 3.1} compared to standard Ensemble Behavior Cloning.
\item \textit{Swarm Behavior Cloning} shows a better performance compared to baseline algorithms in terms of \textit{mean episode return}.
\end{itemize}
\ \\
For testing these hypotheses we used a large set of eight different OpenAI Gym environments \cite{brockman2016openai}: \textit{HalfCheetah, BipedalWalker, LunarLander-continuous, CartPole, Walker2D, Hopper, Acrobot and Ant}. They resemble a large and diverse set of environments containing discrete and continuous action spaces and observation space sizes ranging from $4$-dim to $27$-dim.

\begin{figure*}[h]
\centering
\minipage{1\textwidth}
  \centering
  \includegraphics[width=\linewidth]{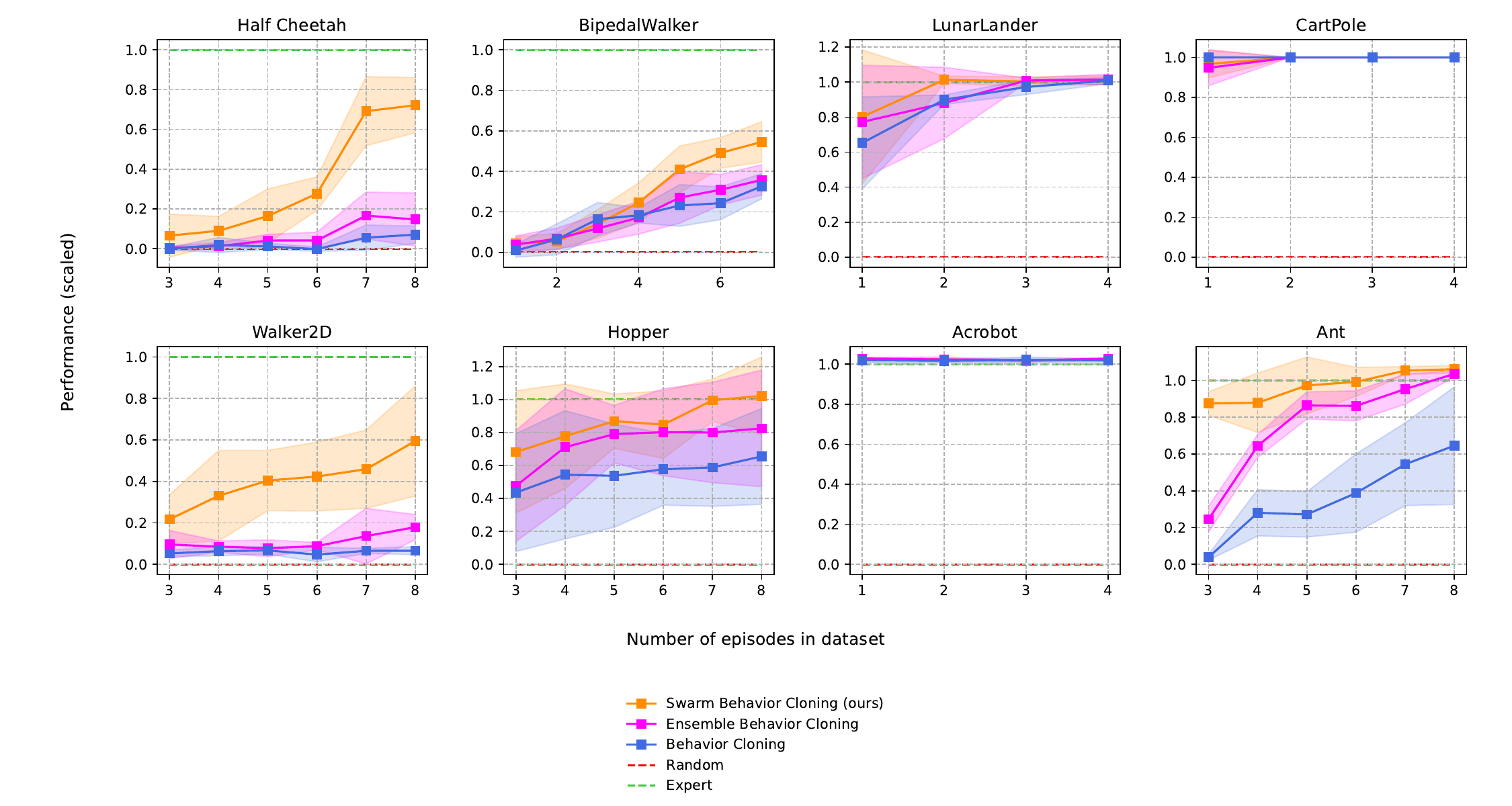}
  \caption{These plots show the mean normalized test returns of our approach \textit{Swarm BC} and two baseline algorithms on eight different OpenAI Gym environments. The graphs represent the mean over 20 episodes and 5 seeds. The x-axes represent the number of expert episodes in the training data $D$. The results show a significant performance improvement in environments with larger observation- and action spaces.}
\endminipage
\end{figure*}
To examine if \textit{Swarm BC} improves the test performance of the agent we used a similar setting as in \cite{ho2016generative}. We used trained SAC- (for continuous action spaces) and PPO- (for discrete action spaces) models from Stable-Baselines 3 \cite{stable-baselines3} as experts and used them to create datasets $D$ containing $x \in [1,8]$ episodes.
Then we trained our approach and two baseline approaches until convergence. We repeated this procedure for 5 seeds. The result is presented in Figure 3.

For easier comparison between environments, we scaled the return. For this, we first determined the mean episode return following the expert policy $R^{expert}$ and the random policy $R^{random}$. Then we used the formula $R^{scaled} = (R - R^{random}) / (R^{expert} - R^{random})$ for scaling the return into the interval [0,1]. 0 represents the performance of the random policy and 1 of the expert policy.

The solid lines in Figure 3 show the mean test performance for 20 episodes and 5 seeds. The shaded areas represent the standard deviation.

The main conclusion we can draw from this experiment is that \textit{Swarm BC} was nearly never worse than \textit{BC} or \textit{Ensemble BC} and in larger environments significantly better. In \textit{HalfCheetah}, for example, the agent achieved a mean scaled episode return of $0.72$ using \textit{Swarm BC} for datasets $D$ containing $8$ expert episodes and just $0.17$ using \textit{Ensemble BC}. \textit{Ensemble BC} still performed better than \textit{single BC} in most environments.
\ \\
\ \\
To test whether our approach does reduce the \textit{mean action difference} as introduced in \textbf{Definition 3.1} we used the same trained models from the previous experiment and calculated the \textit{mean action difference} for each timestep in the test episodes. The x-axes in Figure 4 represent the timestep and the y-axes represent the \textit{mean action difference}. The plots show the average for 20 episodes and 5 seeds. \textit{Swarm Behavior Cloning} did reduce the \textit{mean action difference}, but not always to the same extend. In the \textit{BipedalWalker} environment it was reduced by almost $44\%$ while in \textit{Ant} it was only reduced by $11\%$.

Nevertheless, we can verify the hypothesis that \textit{Swarm BC} does reduce the \textit{mean action difference}.

\begin{figure}[h]
\includegraphics[width=\linewidth]{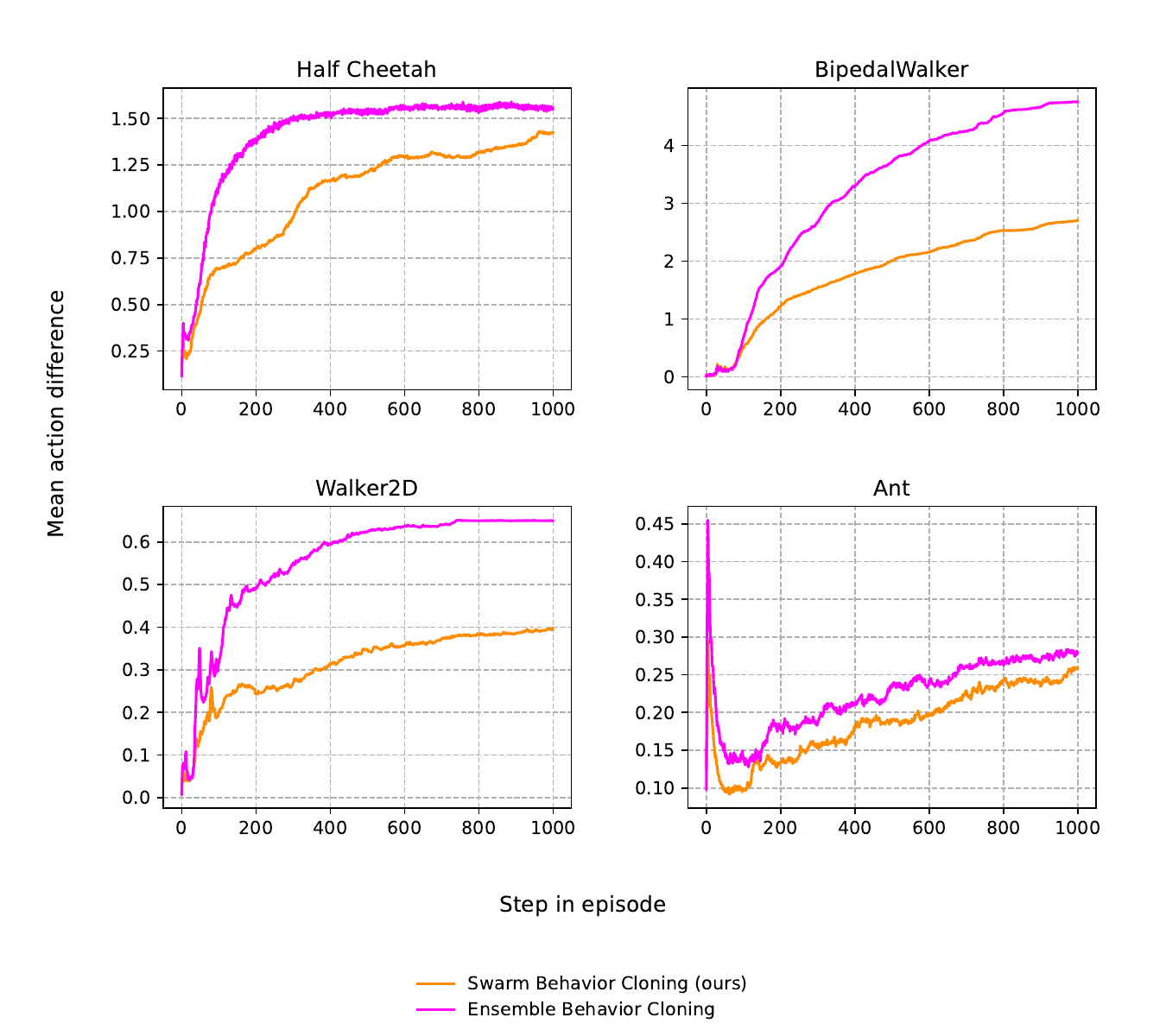}
  \caption{In this figure we are evaluating whether \textit{Swarm BC} can reduce the \textit{mean action difference} as defined in Definition 3.1, which is the difference between the $N$ predicted actions $\{a_i = \pi_i(s)\}_{1 \leq i \leq N}$ of an ensemble $E$ containing $N$ policies. The results show that our approach does indeed reduce it but depending on the environment not always to the same extent. The x-axes in these plots represent the timestep in the test episodes and the y-axes represent the \textit{mean action difference}. The graphs are the mean over $20$ episodes and $5$ seeds.}
\end{figure}
\ \\
\ \\
\textit{Swarm BC} shows significantly better performance compared to baseline algorithms with nearly no computational overhead. The main disadvantage however is the introduction of another hyperparameter $\tau$. To test the sensitivity of this parameter we conducted an ablation study regarding $\tau$ and also about $N$ (the number of policies in the ensemble $E$). The results are plotted in Figure 5. For this ablation study, we used the \textit{Walker2D} environment and again scaled the return for better comparison. For $\tau$ we tested values within $\{0.0, 0.25, 0.5, 0.75, 1.0\}$. The main conclusion for the ablation in $\tau$ is that too large values can decrease the performance. $\tau = 0.25$ worked best so we used this value for all other experiments in this paper.

For the ablation on $N$ we tested values in the set $\{2, 4, 6, 8\}$. For $N = 2$ the test performance was significantly below the test performance for $N = 4$. For larger ensembles ($N = 6$ and $N = 8$) the performance did not increase significantly anymore. But since the training time scales linearly with the number of policies $N$ we chose $N = 4$ for all experiments.

\begin{figure}[h]
\includegraphics[width=\linewidth]{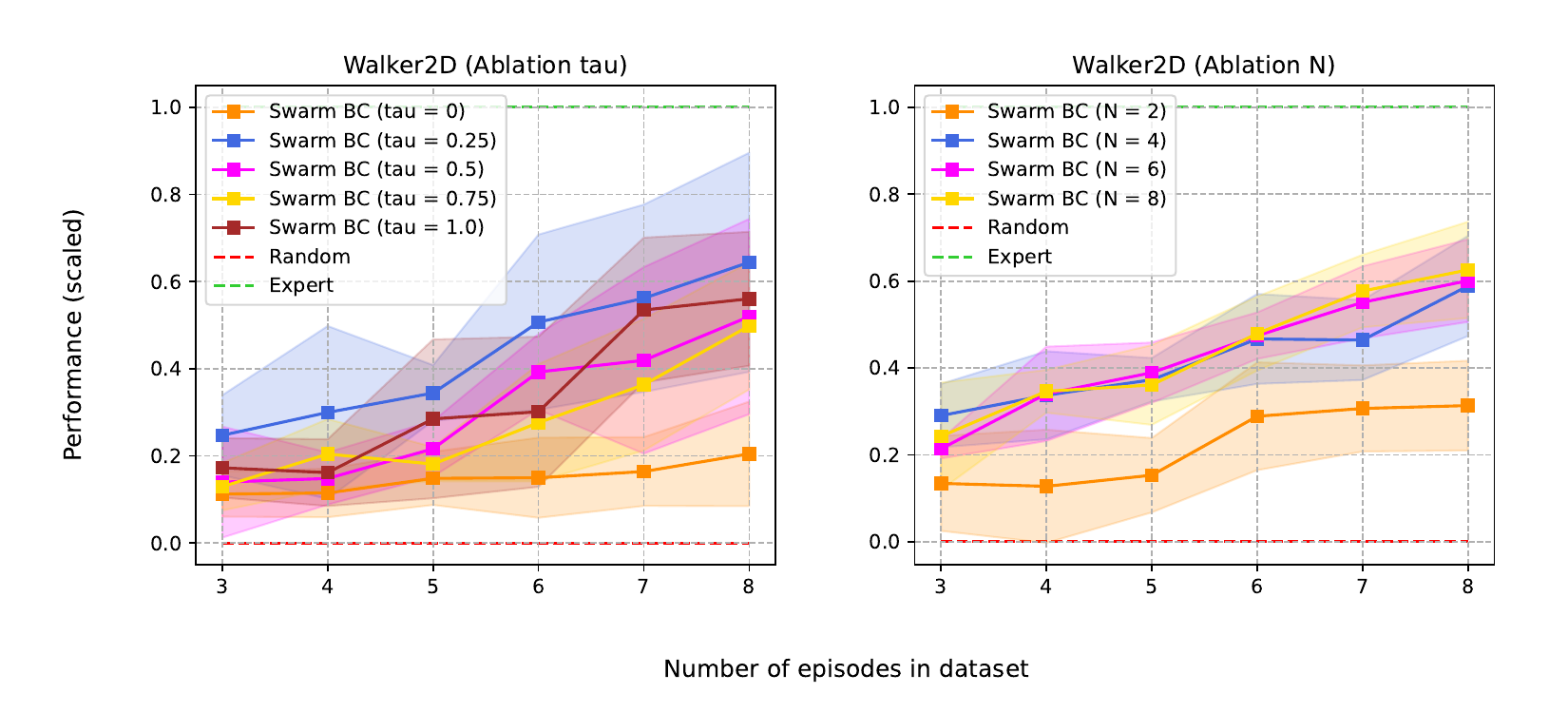}
  \caption{To examine the sensitivity of the two hyperparameters $\tau$ and $N$ we did an ablation study. (Left) choosing $\tau$ too large or too small can reduce the test performance in terms of \textit{mean episode return}. For Walker2D the best value was $\tau = 0.25$. Thus, we chose this value for all experiments in this paper. (Right) the conclusion of the ablation on the ensemble size $N$ is that larger $N$ are better, but this comes at the expense of longer runtime. For $N > 4$, however, the performance does not increase significantly anymore. Thus we chose $N = 4$ for all experiments in this paper.}
\end{figure}
\ \\
As a conclusion of this experiments section we can verify both hypotheses that \textit{Swarm BC} increases test performance and decreases the \textit{mean action difference}. In the next chapter, we provide a theoretical analysis of our approach.

\section{\uppercase{Theoretical Analysis}}

Let $(s,a) \in D$ be a random but fixed state-action tuple out of the training dataset $D$. Let $E = \{\pi_i\}_{1 \leq i \leq N}$ be an ensemble of $N$ policies, each represented as an MLP containing $K$ hidden layers. For state $s$ the ensemble $E$ produces $N$ hidden feature activation $h_{i,k}$ for each layer $k \in [1,K]$.

The basic idea of our approach \textit{Swarm BC} is to train an ensemble $E$ that produces similar hidden features:
\[ \forall \; (i,j) \in [1,N]^2, k \in [1,K]: \;\; h_{i,k} \approx h_{j,k} \]
By doing so the ensemble tries to find features $h_{i,k}$ that can be transformed by at least $N$ different transformations to the desired action $a$ since we have no restriction regarding the weights $W, b$:
\[ h_{i,k+1} = \sigma (W_{i,k} \cdot h_{i,k} + b_{i,k}) \]
Training a single neural net $f_{\phi}$ with parameters $\phi$ on $D$ with some fixed hyperparameters $Q$ corresponds to sampling from the probability density function (pdf):
\[ \phi \sim p(\phi \; | \; D, Q) \]
Since $D$ and $Q$ are fixed we can shorten this expression to $p(\phi)$. The pdf for hidden features $h_k$ for state $s$ corresponds to the integral over all weights that produce the same feature activations:
\[ p(h_k \; | \; s) = \int_{\phi} \mathds{1} [\hat{f^k_{\phi}}(s) = h_k] \cdot p(\phi) \]
For a fixed state $s$ we can shorten this expression to $p(h_k)$.
\ \\
\ \\
\textbf{We now show that training an ensemble with similar feature activations corresponds to finding the \textit{global mode} of the pdf $p(h_k)$}.
\ \\
\ \\
If we are training a standard ensemble, we are sampling $N$ times independently from the pdf $p(h_k)$. But the pdf for sampling $N$ times the same hidden feature activation corresponds to:
\[ p^N(h_k) = \frac{{p(h_k)}^N}{\int_{\varphi} {p(\varphi)}^N} \]
with
\[ {p(h_k)}^N = \prod_{i = 1}^{N} p(h_k) \]
For $N \rightarrow \infty$ the pdf $p^N(h_k)$ corresponds to the Dirac delta function being $p^N(h_k)=+\infty$ for the mode $h_k^+$ of $p(h_k)$ and 0 elsewhere (if there is just one mode). So we just need to sample once from $p^N(h_k)$ to get the mode $h_k^+$. Note that the \textit{probability density} $p(h_k)$ is not the \textit{probability} for sampling $h_k$. The probability for sampling a specific $h_k$ is always 0. The probability can just be inferred by integrating the density over some space. We use as a space the hypercube $T$ of length $\tau$ around activation $h_k$:

\[ P^N_{\tau}(h_k) = \int_{T} p^N(h_k) \]
\ \\
\textbf{\textit{Proposition:}} For $\tau \rightarrow 0$ and $N \rightarrow \infty$, the probability for sampling the global mode with maximum error $\tau$ from $p^N(h_k)$ is $P^N_{\tau}(h_k^+) = 1$ if $p(h_k)$ is continuously differentiable, there is just a single mode $h_k^+$ and the activation space $H_k \ni h_k$ is a bounded hypercube.
\ \\
\ \\
\textit{Proof}: By assumption we know that the activation space $H_k$ is a bounded hypercube of edge length $l$ and number of dimensions $m$. We further know that $h_k^+$ is the only mode of $p(h_k)$ and $p(h_k)$ is continuously differentiable (i.e. $p(h_k)$ is differentiable and its deviation is continuous which implies that there is a maximum absolute gradient).

For a given $\tau \in (0;\infty)$ we split the hypercube $H_k$ in each dimension into $\bigl \lceil \frac{l}{\tau} \bigr \rceil$ parts. Thus $H_k$ is split into ${\bigl \lceil \frac{l}{\tau} \bigr \rceil}^m$ sub-hypercubes. Each of them has maximum volume $\tau^m$. If the mode $h_k^+$ lies on the edge between two sub-hypercubes we move all sub-hypercubes by $\tau/2$. So we need a maximum of  ${\bigl \lceil \frac{l}{\tau} + 1 \bigr \rceil}^m$ sub-hypercubes to ensure that $h_k^+$ lies in exactly one sub-hypercube. We name it $H^+$ and all other sub-hypercubes are labeled ${\{H_i^-\}}_{1 \leq i < {\bigl \lceil \frac{l}{\tau} + 1 \bigr \rceil}^m}$.
We can calculate the mode for $p(h_k)$ in the remaining space of $H_k$ without $H^+$ as follows:
\[ h_k^{\#} = \underset{h_k \in H_k \backslash H^+}{\mathrm{argmax}} \:\: p(h_k)\]
Now we can calculate the upper bound for the probability mass in each $H^-$ sub-hypercube by integrating the maximal possible density $p(h_k^\#)$ over the maximal possible volume $\tau^m$:
\[ P(H^-) \leq \tau^m \cdot p(h_k^\#) \]
Since $h_k^+$ is the only mode, $H_k$ is bounded and $p(h_k)$ is continuously differentiable there is a $\beta \in \mathbb{R}^+$ such that: $p(h_k^+) = p(h_k^\#) + \beta$. This implies that there is a sub-hypercube $H^*$ inside of $H^+$ with edge length $\tilde{\tau} < \tau$ such that:
\[ \forall \; h_k \in H^*: p(h_k) > p(h_k^\#) + \frac{1}{2} \cdot \Bigl( p(h_k^+) - p(h_k^\#) \Bigr) \]
Thus we can calculate a lower bound for the probability mass in $H^+$:
\[ P(H^+) \geq \tilde{\tau}^m \cdot \biggl[ p(h_k^\#) + \frac{1}{2} \cdot \Bigl( p(h_k^+) - p(h_k^\#) \Bigr) \biggr] =\]\[ = \frac{\tilde{\tau}^m}{2} \Bigl( p(h_k^\#) + p(h_k^+) \Bigr)\]
We can generalize both bounds to $P^{ ^N}(H):$
\[ P^{ ^N}(H^-) \leq \frac{\tau^m \cdot p(h_k^\#)^N}{Z} \]
\[ P^{ ^N}(H^+) \geq \frac{\tilde{\tau}^m \cdot \Bigl( p(h_k^\#)^N + p(h_k^+)^N \Bigr) }{2Z} \]
with $Z$ being the normalization constant:
\[ Z = \int_{h_k \in H_k} {p(h_k)}^N \]
Let $\alpha \in [0;\infty)$ be a threshold. To proof that $P^N_{\tau}(h_k)$ approximates the global mode of $p(h_k)$ for $\tau \rightarrow 0$ and $N \rightarrow \infty$ we need to show that for any $\alpha$ and any $\tau \in (0;\infty)$ we can choose $N \in \mathbb{N}$ such that:
\[ \frac{P^{ ^N}(H^+)}{{\bigl \lceil \frac{l}{\tau} + 1\bigr \rceil}^m \cdot P^{ ^N}(H^-)} \geq \alpha \]
Because this would mean we can shift arbitrarily much probability mass into the sub-hypercube $H^+$ by increasing $N$. For that let's consider the ratio:
\[ \frac{P^{ ^N}(H^+)}{P^{ ^N}(H^-)} \geq \frac{\tilde{\tau}^m \cdot p(h_k^\#)^N + \tilde{\tau}^m \cdot p(h_k^+)^N}{2\tau^m \cdot p(h_k^\#)^N} =\]\[ = \underbrace{\frac{\tilde{\tau}^m}{2\tau^m}}_{\equiv c} \cdot \Biggl( 1 + {\biggl(\frac{p(h_k^+)}{p(h_k^\#)}\biggr)}^N \Biggr)\]
Since $h_k^+$ is the only mode the density $p(h_k^+)$ is larger than $p(h_k^\#)$. We can therefore see that the ratio gets arbitrarily large for $N \rightarrow \infty$. So we can choose $N$ according to:
\[ \frac{P^{ ^N}(H^+)}{{\bigl \lceil \frac{l}{\tau} + 1\bigr \rceil}^m \cdot P^{ ^N}(H^-)} \geq \frac{c}{{\bigl \lceil \frac{l}{\tau} + 1\bigr \rceil}^m} \cdot \Biggl[ 1 + {\biggl(\frac{p(h_k^+)}{p(h_k^\#)}\biggr)}^N \Biggr] \stackrel{!}{\geq} \alpha \]
\[ \Rightarrow N = \Biggl \lceil \frac{ln \Bigl( \frac{{\bigl \lceil \frac{l}{\tau} + 1\bigr \rceil}^m \cdot \alpha}{c} - 1 \Bigr) }{ln \Bigl( \frac{p(\phi^+)}{p(\phi^\#)} \Bigr) } \Biggr \rceil \:\:\:\:\:\:\:\:\:\:\:\: \square \]

\section{\uppercase{Conclusion}}

Behavior Cloning (BC) is a crucial method within Imitation Learning, enabling agents to be trained safely using a dataset of pre-collected state-action pairs provided by an expert. However, when applied in an ensemble framework, BC can suffer from the issue of increasing action differences, particularly in states that are underrepresented in the training data $D = {(s_t, a_t)}_t$. These large \textit{mean action differences} among the ensemble policies can lead to suboptimal aggregated actions, which degrade the overall performance of the agent.

In this paper, we proposed \textit{Swarm Behavior Cloning} (Swarm BC) to address this challenge. By fostering greater alignment among the policies while preserving the diversity of their computations, our approach encourages the ensemble to learn more similar hidden feature representations. This adjustment effectively reduces action prediction divergence, allowing the ensemble to retain its inherent strengths—such as robustness and varied decision-making—while producing more consistent and reliable actions.

We evaluated Swarm BC across eight diverse OpenAI Gym environments, demonstrating that it effectively reduces \textit{mean action differences} and significantly improves the agent's test performance, measured by episode returns.

Finally, we provided a theoretical analysis showing that our method approximates the hidden feature activations with the highest probability density, effectively learning the global mode $h_k^* = \underset{h_k}{\mathrm{argmax}} ; p(h_k ; | ; D)$ based on the training data $D$. This theoretical insight further supports the practical performance gains observed in our experiments.

\bibliographystyle{apalike}
{\small
\bibliography{example}}

\end{document}